\documentclass[sigconf,nonacm]{acmart}

\usepackage{bm}
\usepackage{amsmath} 
\usepackage{xspace}  
\usepackage{graphicx}
\usepackage{booktabs}
\usepackage{multirow}

\usepackage{algorithm}
\usepackage{algpseudocode}

\AtBeginDocument{%
  }

\begin{document}

\newcommand{\ourmethod}{Impactful Bit-Flip Search\xspace}

\title{\ourmethod on Full-precision Models}


\author{Matan Levy}
\affiliation{%
  \institution{Tel Aviv University}
  \city{}
  \country{}}

\author{Nadav Benedek}
\affiliation{%
  \institution{Tel Aviv University}
  \city{}
  \country{}}

\author{Mahmood Sharif}
\affiliation{%
  \institution{Tel Aviv University}
  \city{}
  \country{}}

\renewcommand{\shortauthors}{Trovato et al.}

\begin{abstract}
  Neural networks have shown remarkable performance in various tasks, yet they remain susceptible to subtle changes in their input or model parameters. One particularly impactful vulnerability arises through the Bit-Flip Attack (BFA), where flipping a small number of critical bits in a model’s parameters can severely degrade its performance. A common technique for inducing bit flips in DRAM is the Row-Hammer attack, which exploits frequent uncached memory accesses to alter data. Identifying susceptible bits can be achieved through exhaustive search or progressive layer-by-layer analysis, especially in quantized networks. In this work, we introduce \ourmethod (IBS), a novel method for efficiently pinpointing and flipping critical bits in full-precision networks. Additionally, we propose a Weight-Stealth technique that strategically modifies the model’s parameters in a way that maintains the float values within the original distribution, thereby bypassing simple range checks often used in tamper detection.

\end{abstract}

\maketitle

\section{Introduction}

Neural networks have achieved human level performance across various domains \cite{oord2016wavenet,silver2017mastering, he2015delving} and spurred a growing interest in deploying neural algorithms in real-world. As a result, ensuring the security and robustness of deep neural networks has become a critical concern that must be addressed.

Adversarial attacks \cite{goodfellow2014explaining} present a significant security issue for neural networks, potentially injecting system malfunctions through input noise of constrained magnitude that is imperceptible to humans, but catastrophic for neural networks. In recent years, extensive research has been conducted on both attacking and defending against adversarial examples at the input level of neural networks and also at the network parameters \cite{hong2019terminal}. 

Bit flip attacks \cite{rakin2019bit} on neural networks are a type of fault injection attack where specific bits in the memory storing the neural network's parameters (such as weights and biases) are deliberately altered. These attacks exploit the vulnerability of neural networks to small changes in their parameters, which can lead to significant performance degradation or completely erroneous outputs.

Rowhammer attacks \cite{kim2014flipping} represent a prevalent form of bit-wise software-induced fault attack. This vulnerability enables an attacker to induce single-bit corruption at the DRAM level, making it an ideal candidate for our analysis. The versatility of Rowhammer stems from its minimal requirements: an attacker only needs access to DRAM content, a ubiquitous feature in modern systems. By executing specific memory access patterns, the attacker can exert extreme stress on adjacent memory locations, leading to faults in other stored data, potentially flipping a bit.

In this work we present \ourmethod, a novel method for locating and flipping minimal number of bits in full precision neural models with maximum damage to the model. 

\section{Related Work}

\citet{hong2019terminal} investigated the vulnerability of deep neural networks (DNNs) to hardware fault attacks. These attacks, which induce errors in the hardware executing the DNNs, can lead to significant performance degradation and incorrect outputs. The authors explored how different types of faults, such as bit flips in memory or computational units, impact the accuracy and reliability of DNNs. They found that even small, targeted faults can cause substantial drops in performance, leading to what they term "graceless degradation," where the network's output quality rapidly deteriorates.

This study highlighted the lack of robustness in current DNN architectures against such hardware-induced errors, raising concerns for their deployment in critical applications. The authors proposed various mitigation strategies, including fault detection mechanisms and fault-tolerant design techniques, to improve the resilience of DNNs. 

To find the important bits, the authors flipped each bit in all parameters of a \textbf{32-bit single precision float} model, in both directions---(0$\rightarrow$1) and (1$\rightarrow$0)---and compute the RAD over the entire validation set. This \textbf{exhaustive} search can be costly when the parameter size of a model is large. 

To quantify the damage of a single bit-flip, the authors defined the Relative Accuracy Drop as:
$$
RAD = \frac{A_{pristine} - A_{corrupted}}{A_{pristine}}
$$

Where $ A_{pristine} $ 
and $ A_{corrupted} $
denote the classification accuracies of the pristine and the corrupted models, respectively.

$\\$
\citet{bai2023versatile} explores a new type of adversarial attack on deep neural 32-bit float networks (DNNs) that targets the network's weights by flipping a limited number of bits. The authors demonstrated that by strategically flipping just a few bits in the binary representation of the weights, they can significantly degrade the performance of the DNN, leading to incorrect outputs and reduced accuracy. However, they require solving complex mixed integer optimization problems.
$\\$
\citet{rakin2019bit} introduced a Progressive Bit Search (PBS) method for attacking neural networks by progressively searching for and flipping specific bits in the network's weights. The key novelty of this approach lies in its efficiency and effectiveness in degrading neural network performance with minimal bit modifications. Instead of randomly flipping bits or targeting all bits equally, this method systematically evaluates the impact of flipping each bit and prioritizes those that cause the most significant performance degradation, hence reducing the computational cost of the attack and making it harder to detect compared to more extensive weight modifications.

In \cite{rakin2019bit}, for each attack iteration, the process of progressive bit searching is generally divided into two successive
steps:

\begin{enumerate}
    \item In-layer Search: the in-layer search is performed
through selecting the most vulnerable bits in the each
layer, one at a time, then record the inference loss if those bits are
flipped. 
    
    \item  Cross-layer Search: After the in-layer search
conducted upon each layer of the network independently,
the cross-layer search evaluates the recorded loss increment caused by Bit-Flip Attack (BFA) with in-layer search, to identify the
top vulnerable bits across the different layers, and then flip those bits in that layer. The default number of bits that are flipped every iteration is 1. The process may continue for additional in-layer searches as needed.
\end{enumerate}

However, the PBS hardware attack method assumes quantized neural
networks with integer weights, limiting its applicability to a narrow set of models.

\section{Technical Approach}
Our research efforts are focused at finding both theoretically and empirically more efficient and effective methods at locating the critical bits, rather than conducting exhaustive scanning of bits.

Our novel approach stems from the mathematical properties of the relationship between the loss function of the model to the contributing of individual bits under the floating points representations.

The loss contribution of a single bit $b$ residing in a specific float weight $w$ can be phrased using the chain rule:

\begin{equation}
 \frac{\partial L}{\partial b} = \frac{\partial L}{\partial w} \frac{\partial w}{\partial b}
  \label{eq:chain}
\end{equation}

A naive heuristics can first identify those float weights which affect the loss the most and then identify the bits that affect the float the most. However, this approach is susceptible to situations in which a float is not the most impactful among floats, yet a bit in it has a large float-to-bit gradient, resulting in an overall highest loss-to-bit gradient.

Previous work \cite{rakin2019bit} has shown that most frequently the bit that causes the highest RAD (Relative Accuracy Drop) is the bit corresponding to the most significant bit in the exponent part of the float.

A 32-bit big-endian floating point can be represented as follows:

\begin{equation}
  w = \underbrace{(-1)^{b_0}}_{\text{sign}} \underbrace{2^{-127+\sum_{i=0}^{7} 2^{7-i}b_{i+1}}}_{\text{exponent}}\underbrace{(1+\sum_{i=1}^{23}2^{-i}b_{i+8})}_{\text{mantissa}}
  \label{eq:a}
\end{equation}

To find the bits that maximize the inference loss of a model, let us differentiate a float $w$ by the $b_1$, the MSB of the exponent component:

\begin{equation}
  \frac{\partial w}{\partial b_1}=(-1)^{b_0} 2^{-127+\sum_{i=0}^{7} 2^{7-i}b_{i+1}}  (ln(2)2^7)(1+\sum_{i=1}^{23}2^{-i}b_{i+8})=ln(2)2^7 w
\end{equation}

Applying this to the chain rule in Equation~\ref{eq:chain} to obtain:

\begin{equation}
  \frac{\partial L}{\partial b_1}=\frac{\partial L}{\partial w} ln(2)2^7 w
  \label{eq:losstomsb}
\end{equation}

To find the most impactful bit $\hat{b_1}$with the highest absolute gradient with the respect to the loss, denote the vector $\bm{w}$ of all float weights of a model, and then:

\begin{equation}
  \hat{b_1} = argmax_{i} (\lvert \bm{w} \circ \frac{\partial L}{\partial \bm{w}} \rvert)_i 
  \label{eq:mostimpactful}
\end{equation}

Hence searching the most impactful $b_1$ bit can be done efficiently by computing the element-wise absolute value of the element-wise product between the weights' gradient multiplied by the weights, and finding the maximum entry.

Once an impactful bit is detected, is can be flipped if the bit is 0 and $w \frac{\partial L}{\partial w}>0$ or if it is 1 and $w \frac{\partial L}{\partial w}<0$.

Similarly, \eqref{eq:losstomsb} can be generalized to apply to other bits in the exponent of a weight. Specifically, the gradient of the \textit{i}-th bit of the exponent is:

\begin{equation}
  \frac{\partial L}{\partial b_i}=\frac{\partial L}{\partial w} ln(2)2^{7-i} w
  \label{eq:losstoexponentbit} 
\end{equation}

Hence, we can apply \eqref{eq:mostimpactful} for other $\hat{b_i}$'s, which will be useful for algorithm \ref{alg:weightstealth}.

\subsection{Weight-stealth}
Is it possible to compromise a model while keeping its weights within the original distribution to evade detection? In this paper, we investigate the concept of weight-stealth bit flipping, where the model is disrupted yet the altered weights remain within the original minimum-maximum range of the weights. This helps to avoid detection by minimizing conspicuous changes or anomalies in the weight histogram, which might otherwise indicate tampering. Algorithm \ref{alg:weightstealth} outlines the method.

\section{Method}
\label{method}

\subsection{Baseline methods}

The baseline methods for comparison: \newline

\textbf{Random Bit Flip}: A predefined number of randomly selected MSB in the model weights are being flipped. \newline

\textbf{Exhaustive Search}: Algorithm \ref{alg:exhaustive} describes the method.

\begin{algorithm}
\caption{Exhaustive Search \label{alg:exhaustive}}
\begin{algorithmic}[1]
\For {each weight in the model}

    \State Flip the MSB

    \State Compute the validation-batch loss, using a forward pass

    \State Restore the bit value

\EndFor

\State Flip the $n_b$ bits with the highest calculated loss
\end{algorithmic}
\end{algorithm}

\subsection{Our methods}

We analyze the following methods: \newline

\textbf{Model-wise}: 
The objective is to compromise the model's functionality by maximizing the RAD, while flipping as few bits as possible. All model weights are treated collectively as a single block of floats, without regard to their association with individual layers. Only MSB bits are targeted for flipping.
Algorithm \ref{alg:model_wise} describes the method.

\begin{algorithm}
\caption{Model-wise bit flipping \label{alg:model_wise}}
\begin{algorithmic}[1]

\State Using a validation-set batch, make a forward and backward pass, to computer the gradients of the model weights

\State For all the MSB, calculate \eqref{eq:losstomsb}, and sort, to find the most promising bits

\State Filter out the most promising $n_b$ bits, that would lead to an increase in validation-batch loss if flipped: if the bit is 0 and $w \frac{\partial L}{\partial w}>0$ or if it is 1 and $w \frac{\partial L}{\partial w}<0$

\State Flip the filtered bits simultaneously

\end{algorithmic}
\end{algorithm}

\textbf{Layer-wise}: Algorithm \ref{alg:layerwise} describes the method. The objective is similar to that of the model-wise algorithm. In contrast to \citet{rakin2019bit}, we do not use an additional outer loop. This is because, once we flip the MSB bits of the weights, the gradients of the other weights become zero.\newline

\begin{algorithm}
\caption{Layer-wise bit flipping algorithm\label{alg:layerwise}}
\begin{algorithmic}[1]
\For {each layer}

    \State Using a validation-set batch, make a forward and backward pass, to compute the gradients of the model weights

    \State For all the MSB in the current layer, calculate \eqref{eq:losstomsb}, and sort, to find the most promising bits

    \State Filter out the most promising $n_b$ bits, that would lead to an increase in validation-batch loss if flipped: if the bit is 0 and $w \frac{\partial L}{\partial w}>0$ or if it is 1 and $w \frac{\partial L}{\partial w}<0$

    \State Flip the filtered bits simultaneously

    \State Compute the validation-batch loss after the bits are flipped, using a forward pass

    \State Restore the flipped bits

\EndFor

\State Flip bits in the layer with the maximum loss
\end{algorithmic}
\end{algorithm}

Algorithm \ref{alg:weightstealth} describes the weight-stealth bit-flipping method.

\begin{algorithm}
\caption{Weight-stealth bit flipping \label{alg:weightstealth}}
\begin{algorithmic}[1]

\State Determine the maximum and minimum values for each of the model's weights.

\For {$i$ iterations}

\State Using a validation-set batch, make a forward and backward pass, to compute the gradients of the model weights

\For {each layer}

    \State For all the bits in the current layer, calculate \eqref{eq:losstoexponentbit} while changing the coefficient, depending on the bit position within the float weight, and sort, to find the most promising bits

    \State Filter out the most promising $n_b$ bits, that would lead to an increase in validation-batch loss if flipped: if the bit is 0 and $w \frac{\partial L}{\partial w}>0$ or if it is 1 and $w \frac{\partial L}{\partial w}<0$, with the requirement that the flipped float-value remain in the range calculated at step 1

    \State Flip the filtered bits simultaneously

    \State Compute the validation-batch loss after the bits are flipped, using a forward pass

    \State Restore the flipped bits

\EndFor \: layers

\State Flip $n_b$ bits in the layer with the maximum loss

\EndFor \: iterations

\end{algorithmic}
\end{algorithm}

\section{Experiments}

We evaluated our proposed three attack methods: model-wise, layer-wise, and weight-stealth, against the baseline attacks of random bit flipping and exhaustive search, as described in Section \ref{method}. Each attack was tested with five different configurations of bit flips: 1, 15, 35, 70, and 140, as can be seen in Table \ref{tab:results}. We applied these configurations to two neural models described below: VGG-16 and 50k CNN. Each attack was given access to a batch of samples from the validation set, with a fixed sample size of 128 for our experiments. For each attack and each configuration, we repeated the attack 15 times, using a randomly selected validation set batch for each run (random seed), and computed the mean and standard deviation of the resulting RADs over the entire validation set. The results are summarized in Table \ref{tab:results}. \newline

The following models and datasets were used for the experiments: \newline

\textbf{VGG-19}: Starting with a pretrained VGG-16 model, we adapted the first convolutional layer to handle MNIST's grayscale images. Additionally, we modified the final fully connected layer to match MNIST's class count. The model was then fine-tuned on the MNIST dataset, achieving a validation accuracy of 98.71\%.

\textbf{50k CNN}: A basic CNN 50k weights' network [CRMCRMLRL], where C is a convolutional layer, R is ReLU, M is max-pool, L is a linear layer. Trained for one epoch on the MNIST dataset.

\section{Results \& Discussion}

\begin{table*}[t]
\caption{Comparison of RAD for two models under different bit-flip values. Higher is better.}
\label{tab:results}
\centering
\scriptsize
\setlength{\tabcolsep}{2pt} 
\renewcommand{\arraystretch}{0.85} 
\resizebox{130mm}{!}{ 
\begin{tabular}{cccccc}
\hline
\textbf{Method} & \textbf{1 Bit Flip} & \textbf{15 Bit Flips} & \textbf{35 Bit Flips} & \textbf{70 Bit Flips} & \textbf{140 Bit Flips} \\ \hline
\multicolumn{6}{c}{\textbf{VGG Model}} \\ \hline
Random BF         & $0.00 \pm 0.00$   & $0.13 \pm 0.31$   & $0.12 \pm 0.29$  & $0.22 \pm 0.32$  & $0.61 \pm 0.34$ \\
Model-Wise        & $0.82 \pm 0.04$   & $0.90 \pm 0.00$   & $0.90 \pm 0.00$  & $0.90 \pm 0.00$   & $0.90 \pm 0.00$ \\
Layer-Wise        & $0.82 \pm 0.08$   & $0.89 \pm 0.02$   & $0.84 \pm 0.11$  & $0.90 \pm 0.01$   & $0.90 \pm 0.00$ \\
Weight-Stealth    & $0.00 \pm 0.00$  & $0.19 \pm 0.03$   & $0.29 \pm 0.12$  & $0.84 \pm 0.11$  & $0.88 \pm 0.01$ \\ \hline

\multicolumn{6}{c}{\textbf{50K CNN Model}} \\ \hline
Random BF         & $0.00 \pm 0.00$   & $0.37 \pm 0.43$   & $0.25 \pm 0.39$  & $0.53 \pm 0.38$   & $0.75 \pm 0.17$ \\
Exhaustive Search & $0.88 \pm 0.00$   & $0.96 \pm 0.00$   & $0.97 \pm 0.01$  & $0.93 \pm 0.04$   & $0.89 \pm 0.00$ \\
Model-Wise        & $0.87 \pm 0.03$  & $0.88 \pm 0.03$   & $0.86 \pm 0.04$  & $0.90 \pm 0.04$   & $0.89 \pm 0.00$ \\
Layer-Wise        & $0.88 \pm 0.02$   & $0.89 \pm 0.04$   & $0.86 \pm 0.07$  & $0.87 \pm 0.06$   & $0.87 \pm 0.06$ \\
Weight-Stealth    & $0.00 \pm 0.01$  & $0.22 \pm 0.06$   & $0.61 \pm 0.12$  & $0.86 \pm 0.06$   & $0.90 \pm 0.00$ \\ \hline
\end{tabular}
}
\end{table*}

Examining the \textbf{Exhaustive Search} results for the 50K CNN, we observe that increasing the number of flipped bits beyond 35 leads to a decline in performance. This occurs because each bit-candidate is selected based on the impact of its individual flip; however, when multiple bits are flipped simultaneously, they may produce conflicting effects on the loss. This is due to the local nature of gradients, while even a single bit flip can significantly alter the float value. In all cases, as one would expect, Exhaustive Search outperforms Random BF. Notably, the exhaustive search baseline was not applied to the VGG-16 model due to the high computational cost associated with its large parameter count, making this approach impractical.\newline

The results reveal that both model-wise and layer-wise attacks are highly effective, with a single bit flip being sufficient to reduce the accuracy of both VGG-16 and the 50K CNN model to negligible levels. Additionally, these two attacks require only a few MSB bit flips to achieve an RAD near 0.9, effectively reducing model performance to that of random guessing. This underscores the efficiency and power of model-wise and layer-wise attacks, as they approach the effectiveness of exhaustive search while being far more computationally efficient.  \newline

Moreover, table \ref{tab:results} demonstrate that there is no advantage to the layer-wise method (in the non weight-stealth setting), and one can treat all model weights as a single chunk. \newline

The weight-stealth attack demonstrates a distinct behavior: its RAD correlates with the number of bit flips. Importantly, as the number of bit flips increases, it achieves similar RAD levels to those of the model-wise, layer-wise, and exhaustive search attacks. This establishes the weight-stealth attack as a novel, powerful approach for reducing model performance to chance levels while remaining unobtrusive by preserving the overall weight distribution.  \newline

\textbf{Weight-Stealth Attack Configurations}
Given the iterative nature of the weight-stealth algorithm, it can be configured in various ways. Table \ref{tab:weight_stealth_attack_RAD} compares the performance of the weight-stealth attack under different configurations, each flipping a total of 70 bits. Configurations are defined by the number of iterations and the number of bits flipped per iteration. The results suggest that increasing the number of iterations improves attack efficacy, likely due to the additional conditioning of each bit flip on prior iterations. However, increasing iterations also extends runtime, resulting in a trade-off between RAD effectiveness and computational cost. \newline

\begin{figure}[h]
    \centering
    \includegraphics[width=0.5\textwidth]{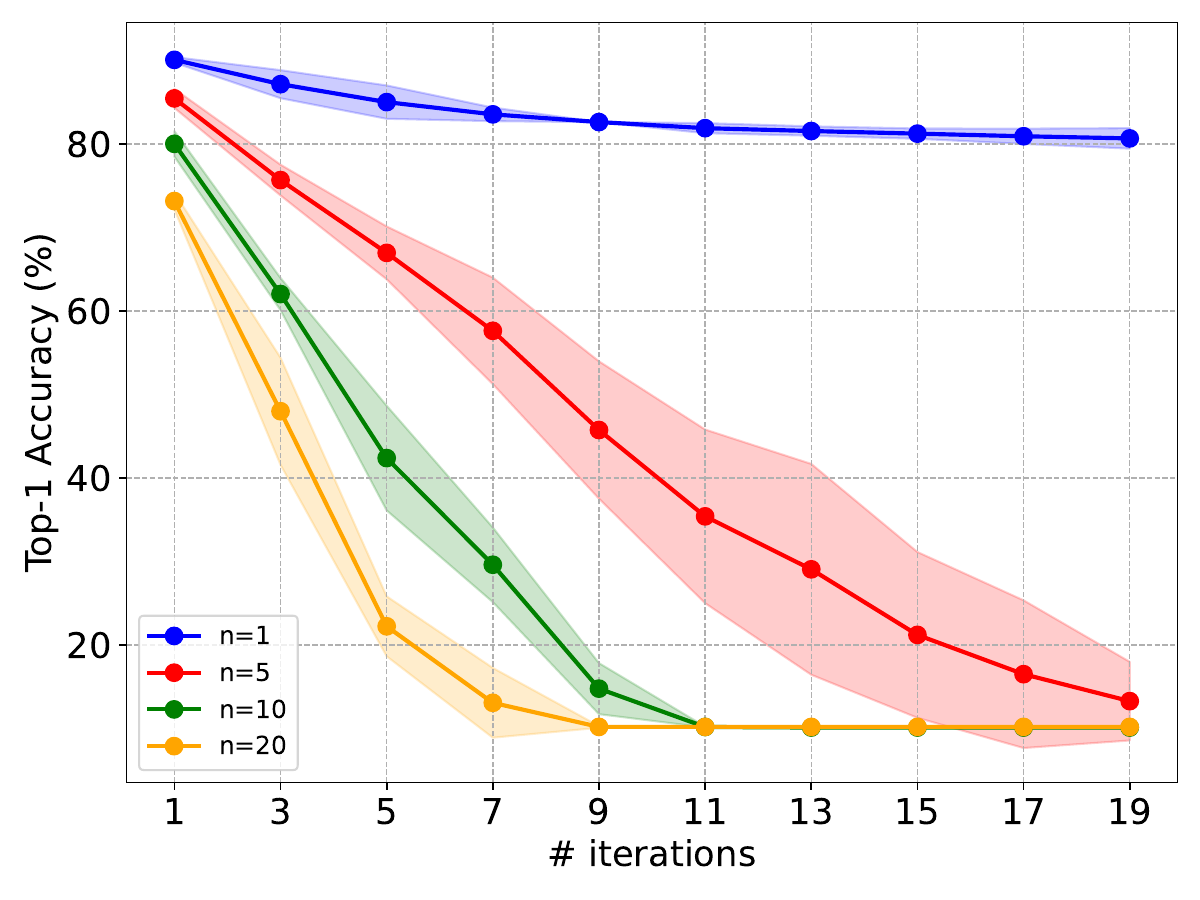} 
    \caption{Post attack accuracy comparison between Weight-Stealth attacks with different number of bit flips per iterations. $n$ denotes the number of bit flips per iteration. The target model is the 50K CNN. Lower is better. Shaded areas are standard deviations.}
    \label{fig:WSnvsk}
\end{figure}

Figure \ref{fig:WSnvsk} further illustrates this trade-off by showing the effect of fixing the number of bit flips per iteration and varying the number of iterations. The results indicate that while more bit flips per iteration accelerate the degradation of model accuracy, weight-stealth configurations with a higher number of iterations yield better performance, given the same total number of bit flips.

\begin{table}[htbp]
\centering
\caption{Comparison of Weight-Stealth attack for different configurations on two target models. The results represent RAD. The attacks were performed with a sample size of 256. $k$ denotes the number of iterations, and $n_b$ is the number of bit flips per iteration. Higher is better.}
\label{tab:WSconfig}
\begin{tabular}{|c|c|c|}
\hline
$(k, n_b)$ & \textbf{50K CNN (RAD) $\uparrow$} & \textbf{VGG (RAD) $\uparrow$} \\ \hline
\textbf{(1, 70)} & 0.69 $\pm$ 0.11 & 0.08 $\pm$ 0.01 \\ \hline
\textbf{(2, 35)} & 0.79 $\pm$ 0.10 & 0.35 $\pm$ 0.23 \\ \hline
\textbf{(5, 14)} & 0.88 $\pm$ 0.02 & 0.67 $\pm$ 0.15 \\ \hline
\textbf{(10, 7)} & 0.89 $\pm$ 0.02 & 0.84 $\pm$ 0.05 \\ \hline
\textbf{(14, 5)} & 0.88 $\pm$ 0.04 & 0.79 $\pm$ 0.12 \\ \hline
\end{tabular}
\label{tab:weight_stealth_attack_RAD}
\end{table}

\section{Conclusion}

In this paper, we presented three approaches for compromising float-weight models: two that are unconstrained and one that is constrained to keep the float-weights within their natural distribution to avoid detection.

We demonstrated a clear advantage over both random and exhaustive methods, and also established that layer-wise processing is unnecessary.

The weight-stealth method we presented can effectively disable a model, reaching approximately 90\% RAD with just 140 bit flips per each of the two models, all while keeping the float-weights within their original range.

Recognizing that in the non-weight-stealth setting, layer-by-layer processing provides no clear advantage, a natural follow-up question is how the weight-stealth algorithm would perform if all model weights were treated as a single block, similar to the approach used in the non-weight-stealth experiments. We leave this exploration for future work.

\section{Contributions}

Levy, Benedek: contributed equally to this work. 
Sharif: Supervision.

\bibliographystyle{ACM-Reference-Format}
\bibliography{sample-base}

\end{document}